\documentclass[letterpaper, 10 pt, conference]{ieeeconf}  %

\IEEEoverridecommandlockouts                              %

\usepackage{mathtools,amssymb}
\usepackage{amsmath}
\usepackage{graphicx,import}
\usepackage{verbatim}
\usepackage{siunitx}
\usepackage{cite}
\usepackage{hyperref}
\hypersetup{
    colorlinks=false,
    citecolor=[rgb]{0,0,0},
    linkcolor=[rgb]{0.1,0.1,0.1},
    urlcolor=[rgb]{0.0235,0.2706,0.6784},
}
\usepackage{cleveref}
\usepackage{microtype}
\usepackage[yyyymmdd]{datetime}
\usepackage{subcaption}
\usepackage{ragged2e}
\usepackage[table]{xcolor}

\usepackage{tikz}
\usetikzlibrary{calc}
\newcommand*\circled[1]{\tikz[baseline=(char.base)]{
    \node[shape=circle, draw, inner sep=1pt, 
        minimum height={\f@size*1.6},] (char) {\vphantom{WAH1g}#1};}}

\usepackage{soul}
\newcommand\change[1]{#1}

\usetikzlibrary{intersections}
\usepackage{circuitikz}
\usepackage{booktabs}

\usepackage{amsthm} %
\theoremstyle{definition}
\theoremstyle{definition}
\theoremstyle{definition}
\theoremstyle{definition}\newtheorem{assumption}{Assumption}
\theoremstyle{definition}
\theoremstyle{definition}
\theoremstyle{definition}
\theoremstyle{definition}
\theoremstyle{definition}
\theoremstyle{definition}
\theoremstyle{remark}

\newdate{datetimeURL}{30}{6}{2023}

\newcommand{\brRound}[1]{\left( #1 \right)}             %
\newcommand{\brSquare}[1]{\left[ #1 \right]}            %
\newcommand{\brCurly}[1]{\{ #1 \}}                      %
\newcommand{\bmat}[1]{\begin{bmatrix} #1 \end{bmatrix}}   %
\newcommand{\bmatSmall}[1]{\begin{bsmallmatrix} #1 \end{bsmallmatrix}}

\newcommand{\R}[1]{\mathbb{R}^{#1}}                     %
\newcommand{\SO}[1]{\text{SO} {\brRound{#1}}}             %

\newcommand{\eye}[1]{{I}_{#1}}                          %
\newcommand{\zero}[1]{\boldsymbol{0}_{#1}}              %
\newcommand{\tr}{\top}                                  %

\newcommand{\frm}[1]{\mathcal{#1}}                      %
\newcommand{\frmWorld}{\frm{I}}                         %
\newcommand{\frmBase}{\frm{B}}

\newcommand{\rotm}[2]{R^{#1}_{#2}}                          %
\newcommand{\rotmW}[1]{\rotm{}{#1}}                         %
\newcommand{\quat}[2]{\boldsymbol{\rho}^{#1}_{#2}}                %
\newcommand{\quatW}[1]{\quat{}{#1}}                         %
\newcommand{\posW}[1]{\boldsymbol{p}_{#1}}                  %
\newcommand{\linVelW}[1]{\boldsymbol{\dot{p}}_{#1}}         %
\newcommand{\angVelW}[1]{\boldsymbol{\omega}_{#1}}          %
\newcommand{\linAccW}[1]{\boldsymbol{\ddot{p}}_{#1}}        %
\newcommand{\angAccW}[1]{\boldsymbol{\dot{\omega}}_{#1}}         %
\newcommand{\versorW}[2]{\boldsymbol{\hat{#1}}_{#2}}             %
\newcommand{\FK}[1]{\text{FK}_{#1}}                          %

\newcommand{\forceBody}[2]{\boldsymbol{f}^{#1}_{#2}}
\newcommand{\torqueBody}[2]{\boldsymbol{\mu}^{#1}_{#2}}

\newcommand{\configurationRobot}{\boldsymbol{q}}
\newcommand{\configurationSpaceRobot}{\mathbb{Q}}
\newcommand{\configurationSpaceVelocityRobot}{\mathbb{V}}
\newcommand{\nuRobot}{\boldsymbol{\nu}}
\newcommand{\dotNuRobot}{\boldsymbol{\dot{\nu}}}
\newcommand{\massMatrixRobot}{M}

\newcommand{\biasVectorRobot}{\boldsymbol{h}}

\newcommand{\jacobianRobot}[2]{J^{#1}_{#2}}

\newcommand{\jointTorRobot}{\boldsymbol{\tau}}
\newcommand{\jointDotTorRobot}{\boldsymbol{\dot{\tau}}}
\newcommand{\jointPosRobot}{\boldsymbol{s}}
\newcommand{\jointVelRobot}{\boldsymbol{\dot{s}}}
\newcommand{\jointAccRobot}{\boldsymbol{\ddot{s}}}
\newcommand{\propThrustRobot}{u}
\newcommand{\propDotThrustRobot}{\dot{u}}
\newcommand{\power}{W}

\newcommand{\qvar}{\hspace{-0.5mm}\brRound{\configurationRobot}}
\newcommand{\qnuvar}{\hspace{-0.5mm}\brRound{\configurationRobot, \nuRobot}}

\newcommand{\airDensity}{\varrho_{\text{air}}}

\newcommand{\airSpeed}[1]{\boldsymbol{v_a}^{#1}}

\newcommand{\reynolds}{R \hspace{-0.25mm} e}
\newcommand{\mach}{\mathcal{M}}

\newcommand{\cost}[1]{\mathcal{L}_{#1}}

\title{\bf \LARGE 
\resizebox{\textwidth}{!}{Co-Design Optimisation of Morphing Topology and Control of Winged Drones}
}

\author{Fabio Bergonti$^{1,2}$, \textit{IEEE Member}, Gabriele Nava$^{1}$, \textit{IEEE Member}, Valentin Wüest$^{3}$, \textit{IEEE Member}, \\ Antonello Paolino$^{1,4}$, Giuseppe L'Erario$^{1,2}$, Daniele Pucci$^{1,2}$, \textit{IEEE Member}, and Dario Floreano$^{3}$, \textit{IEEE Fellow}%
\thanks{
$^{1}$Artificial and Mechanical Intelligence Laboratory, Istituto Italiano di Tecnologia (IIT).
$^{2}$School of Computer Science, University of Manchester.
$^{3}$Laboratory of Intelligent Systems, Ecole Polytechnique Federale de Lausanne (EPFL).
$^{4}$Department of Industrial Engineering, University of Naples Federico II. 
This work was partly funded by the European Union's Horizon 2020 research and innovation programme under grant agreement ID: 871479 AERIAL-CORE.
\looseness=-1
}
}

\usepackage{eso-pic}
\usepackage{graphicx} 

\AddToShipoutPictureBG*{%
	\AtPageUpperLeft{%
		\setlength{\unitlength}{1mm}%
		\put(0,-12){\makebox(\paperwidth,0)[c]{\parbox{0.8\textwidth}{\centering\textcolor{gray}{\large This paper has been published in the 2024 International Conference on Robotics and Automation (ICRA), ©IEEE}}}}
	}
}

\begin{document}

\maketitle
\thispagestyle{empty}
\pagestyle{empty}
\everypar{\looseness=-1}
\begin{abstract}

The design and control of winged aircraft and drones is an iterative process aimed at identifying a compromise of mission-specific costs and constraints.
When agility is required, shape-shifting (morphing) drones represent an efficient solution.
However, morphing drones require the addition of actuated joints that increase the topology and control coupling, making the design process more complex.
We propose a co-design optimisation method that assists the engineers by proposing a morphing drone's conceptual design that includes topology, actuation, morphing strategy, and controller parameters.
The method consists of applying multi-objective constraint-based optimisation to a multi-body winged drone with trajectory optimisation to solve the motion intelligence problem under diverse flight mission requirements, \change{such as energy consumption and mission completion time}.
We show that co-designed morphing drones outperform fixed-winged drones in terms of energy efficiency and mission time, suggesting that the proposed co-design method could be a useful addition to the aircraft engineering toolbox.
\looseness=-1

\end{abstract}

\section{Introduction} \label{sec:introduction}

Fixed-wing drones are highly efficient aerial vehicles employed in high-endurance missions, particularly in environments with sparse obstacles that allow ample space for turning~\cite{floreano2015science,kumar2012opportunities}.
Enhancing the manoeuvrability of winged drones is crucial to enable their navigation in complex environments, and one possible improvement is to integrate \textit{morphing wings}~\cite{MR:morphingWingReview2010}. 
Morphing wings can be designed following various strategies, i.e., varying the dihedral angle~\cite{drone:morphing_wing:muav:dihedral:paranjape2013novel}, the wing sweep \cite{drones:bixler3:waldock2018learning}, the angle of incidence \cite{drone:lishack:ajanic2022sharp}, the trailing edge \cite{drone:airfoil:vos2007post}, the span length \cite{vtol:jlion:ke2016systematic}, or twisting the wing \cite{drone:twist:nasa:jenett2017digital}.
Each strategy has its advantages and disadvantages, depending on the desired performance, the application, and the pilot or controller capabilities.
The selection of the morphing strategy is part of the \textit{aircraft design}, which is usually resolved through an iterative process based on heuristics, historical results, and engineer experience~\cite{drones:book:design:raymer2012aircraft}.
Although the design process is discussed in various books~\cite{drones:book:design:raymer2012aircraft,sadraey2012aircraft,drones:book:UAV:keane2017small}, the performances the drone displays depend on the designer's abilities, the economical availability to investigate multiple solutions and perform several iterations with physical prototypes, and, in the case of unmanned vehicles, on the harmonious integration between hardware and control.
Consequently, the designed drones might end up as sub-optimal conventional solutions with control algorithms not optimised for the selected morphing strategies.
To address these issues, in this paper, we propose a \textit{co-design optimisation method} that can assist the engineer in the design phase by proposing \change{optimised} conceptual designs and control strategies for specific mission scenarios.\looseness=-1

Co-design is a %
multidisciplinary approach to %
conceive more efficient robotic platforms by jointly solving hardware and control problems~\cite{CO:hand:chen2021co,CO:garcia2019control}.
This methodology has found successful applications in various robotic systems, including humanoid robots~\cite{CO:sartore2023codesign,CO:sartore2022optimization}, jumping monoped~\cite{CO:monoped:fadini2022simulation}, quadruped~\cite{CO:quadruped:bravo2022large}, robotic hands~\cite{CO:hand:chen2020hardware}, robotic arms~\cite{CO:arm:sathuluri2023robust}, multicopters~\cite{CO:propeller:du2016computational,CO:box:schiano2022reconfigurable,CO:carlone2019robot}, and winged drones~\cite{CO:grammar:zhao2022automatic,CO:flapping:de2007artificial}. %
In the case of flying machines, there exist methods that focus on optimising hardware component selection, but most of them neglect aerodynamics, making them unsuitable for applications involving morphing wings~\cite{CO:propeller:du2016computational,CO:box:schiano2022reconfigurable,CO:carlone2019robot}.
Conversely, \cite{CO:grammar:zhao2022automatic} proposed a framework incorporating aerodynamics but remains limited to vehicles without dynamic shape-shifting surfaces.
\cite{CO:flapping:de2007artificial} has no previous shortcomings but relies on fixed-period joint movements, making it unsuitable for operation in dynamic environments.\looseness=-1

Here, we propose a method for the co-design of multi-bodied winged drones that perform agile and energy-efficient flight trajectories in %
cluttered environments. 
The method consists of multi-objective constraint-based optimisation for a winged drone whose motion strategy is optimised by a trajectory planning algorithm for systematically finding \change{time varying} morphing movements.
The drones are evaluated with a dynamic multi-body modelling that accounts for aerodynamics.
\looseness=-1

\begin{figure}[t]
    \includegraphics[width=\columnwidth]{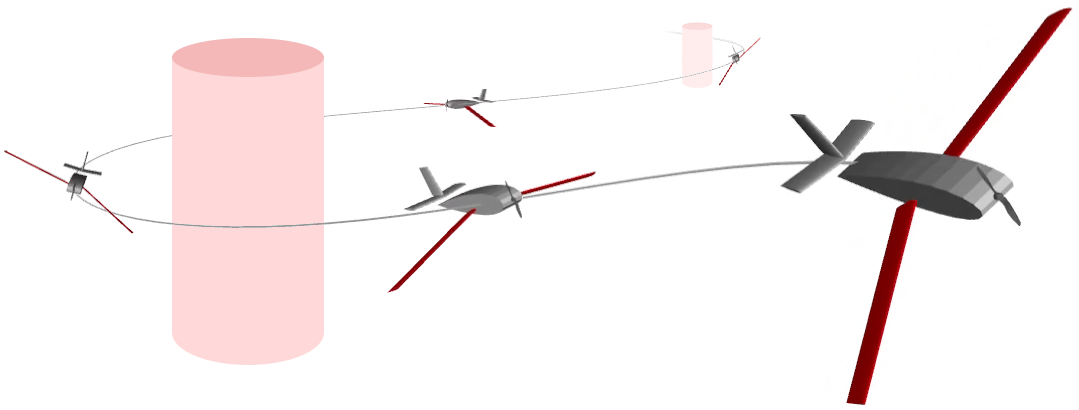}
    \caption{A co-designed morphing drone executing an agile manoeuvre.\looseness=-1
    }
    \label{back:fig:platform_init}
\end{figure}

The rest of this article is organised as follows. 
Section~\ref{sec:background} introduces notation, presents the specificities of the morphing drone considered in this study, and recalls the basis of multibody modelling.
Section~\ref{sec:codesign-framework} introduces the co-design method to evaluate the optimal drone.
Section~\ref{sec:modelling} explains the morphing drone modelling.
Section~\ref{sec:traj-opt} presents the trajectory optimisation problem for solving desired scenarios.
Section~\ref{sec:result} discusses the results obtained with the method and compares the performance of the morphing drone with a commercial fixed-wing drone.
Finally, Section~\ref{sec:conclusions} concludes this article.

\section{Background} \label{sec:background}

\subsection{Notation}

\begin{itemize}
    \item $\eye{n} {\in} \R{n{\times}n}$ identity matrix, $\zero{n {\times}m} {\in} \R{n{\times}m}$ zero matrix.
    \item $\SO{3} := \brCurly{ \rotm{}{} \in \R{3 {\times}3} \; | \; \rotm{\tr}{} \rotm{}{} {=} \eye{3}, \, \det {\brRound{\rotm{}{}}} {=} 1}$.
    \item $\frmWorld$ is the inertial frame; $\frm{C}$ is a generic body-fixed frame.
    \item $\posW{\frm{C}} \in \R{3}$ denotes the origin of frame $\frm{C}$ expressed in $\frmWorld$.
    \item $\rotmW{\frm{C}} \in \SO{3}$ is the rotation matrix that transforms a $3$D vector expressed with the orientation of the frame $\frm{C}$ in a $3$D vector expressed in the frame $\frmWorld$.
    $\rotmW{\frm{C}} {=} \bmat{ \versorW{x}{\frm{C}} & \versorW{y}{\frm{C}} & \versorW{z}{\frm{C}}}$
    with $\versorW{x}{\frm{C}}$, $\versorW{y}{\frm{C}}$, and $\versorW{z}{\frm{C}}$ are the unit vectors of the frame axes.
    \item $\quatW{\frm{C}} \in \R{4}$ is the unit quaternion representation of $\rotmW{\frm{C}}$.
    \item $\angVelW{\frm{C}} \in \R{3}$ is the angular velocity of $\frm{C}$ relative to $\frmWorld$.
\end{itemize}

\subsection{Platform definition} \label{subse:platform}

\begin{figure}[t]
    \resizebox{\columnwidth}{!}{\input{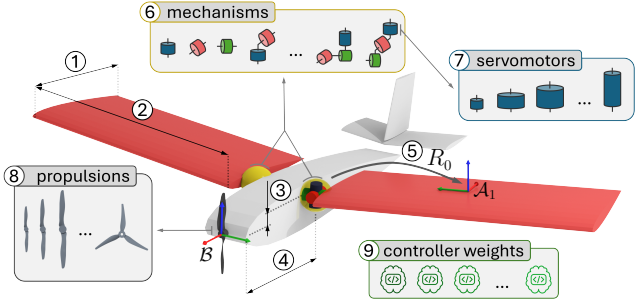}}
    \caption{\change{Morphing drone with its co-design parameters:}
$1\text{-}2)$ \change{wing chord and span size;}
$3\text{-}4)$ \change{wing vertical and horizontal location;}
$5)$ \change{wing static orientations $\rotmW{0}$;}
$6)$ \change{kinematic chain of the morphing mechanism;}
$7)$ \change{servomotor models;}
$8)$ \change{propulsion unit characteristics; and}
$9)$ \change{controller weights. 
}\looseness=-1
    }
    \label{back:fig:platform}
\end{figure}

The platform consists of a winged aircraft with left and right wings symmetrically connected to the fuselage by a \change{mechanism}, which enables relative motion between the parts.
The possible wing movements depend on the type of \change{mechanism} used to connect the wing and the fuselage (see \cref{back:fig:platform}).
For instance, if we implement a \change{mechanism composed of three intersecting revolute joints}, we can obtain sweep, incidence, and dihedral variations.
This solution provides a complete range of motion at the cost of higher weight and energy consumption.
Alternatively, simpler and lighter \change{mechanism} with one or two revolute joints can be used, but it becomes essential to determine the axis of rotation for achieving the desired behaviour.
Although \change{mechanisms} can be designed to produce translation motion, our study focuses on rotational movements only.
Throughout the paper, we refer to this platform with the term \textit{morphing drone}.
\looseness=-1
For the sake of simplicity, our platform does not have wing ailerons, tail rudders, or elevators, leaving only the actuation of the morphing wing to control the drone's attitude.
The propulsion unit consists of an electric propeller positioned at the front of the fuselage (see \cref{back:fig:platform}).
\looseness=-1

\subsection{Floating Multibody System Modelling}

The morphing drone can be modelled as a floating rigid multibody system with $n_{\text{j}}{+}1$ rigid bodies and $n_{\text{j}}$ one degrees-of-freedom (DoF) joints~\cite{featherstone2014rigid}. %
The configuration space $\configurationSpaceRobot$ of the system is the lie group $\configurationSpaceRobot {=} \R{3} {\times} \SO{3} {\times} \R{n_{\text{j}}}$ represented by the pose (i.e. position and orientation) of the \textit{base frame} $\frmBase$, and the \textit{joint positions}. %
The base frame is a frame rigidly connected to the fuselage nose (see \cref{back:fig:platform}).
The configuration $\configurationRobot {=} \brRound{\posW{\frmBase}, \rotmW{\frmBase}, \jointPosRobot}$ is a generic element of the configuration space. $\posW{\frmBase}$ and $\rotmW{\frmBase}$ are the position and orientation of the base frame, and $\jointPosRobot$ are the joint positions.
The tangent space of $\configurationSpaceRobot$ is the configuration space velocity $\configurationSpaceVelocityRobot{=} \R{3} {\times} \R{3} {\times} \R{n_{\text{j}}}$. The configuration velocity $\nuRobot {=} \brRound{\linVelW{\frmBase}, \angVelW{\frmBase}, \jointVelRobot}$ is a generic element of $\configurationSpaceVelocityRobot$. $\linVelW{\frmBase}$ and $\angVelW{\frmBase}$ are the linear and angular velocities of the base frame, and $\jointVelRobot$ are the joint velocities.
The position of a generic frame $\frm{C}$ attached to a link can be computed with the forward kinematic function $\FK{\posW{\frm{C}}} {\brRound{\cdot}}{:} \configurationSpaceRobot {\rightarrow} \R{3}$.
The linear and angular velocity of frame $\frm{C}$ can be computed via the Jacobian $\jacobianRobot{}{\frm{C}} \in \R{6\times6+n_{\text{j}}}$ which maps the configuration velocity to the Cartesian space, namely $\brRound{\linVelW{\frm{C}}; \angVelW{\frm{C}}} = \jacobianRobot{}{\frm{C}} \nuRobot$.
The equation of motion, applying the Euler-Poincaré formalism ~\cite[Ch. 13.5]{marsden2013introduction}, results in
\begin{equation} \label{eq:back:equation_motion}
    \massMatrixRobot \qvar \dotNuRobot + \biasVectorRobot \qnuvar = \bmat{\zero{1\times6} & \jointTorRobot^\tr}^\tr + \boldsymbol{\mathrm{f}}^{\star}.
\end{equation}
$\massMatrixRobot \qvar {\in} \R{n_{\text{j}}{+}6 {\times} n_{\text{j}}{+}6}$ is the mass matrix, 
$\biasVectorRobot \qnuvar {\in} \R{n_{\text{j}}+6}$ is the biased term which accounts for Coriolis, centrifugal effects, \change{friction,} and gravity. $\jointTorRobot {\in} \R{n_{\text{j}}}$ is the vector of joint torques, and $\boldsymbol{\mathrm{f}}^{\star} {\in} \R{n_{\text{j}}+6}$ is the vector of generalised external wrenches.\looseness=-1

\subsection{Wing Aerodynamic Modelling}

A wing immersed in airflow causes deflections that generate stresses on the body. The net effect of the stresses integrated over the complete body surface results in aerodynamic forces and moments, which can be modelled as:
\begin{subequations} \label{eq:aero:force-moments}
\begin{gather} 
    \forceBody{}{\boldsymbol{A}} = \rotmW{\frm{W}} \frac{1}{2} \airDensity \| \airSpeed{} \|^2 S \bmatSmall{ - C_{D} \brRound{\alpha, \beta, \reynolds, \mach} \\ C_{Y} \brRound{\alpha, \beta, \reynolds, \mach}  \\ - C_{L} \brRound{\alpha, \beta, \reynolds, \mach} }, \\
    \torqueBody{}{\boldsymbol{A}} = \rotmW{\frm{W}} \frac{1}{2} \airDensity \| \airSpeed{} \|^2 S \bmatSmall{ b \, C_{l} \brRound{\alpha, \beta, \reynolds, \mach} \\ c \, C_{m} \brRound{\alpha, \beta, \reynolds, \mach}  \\ b \, C_{n} \brRound{\alpha, \beta, \reynolds, \mach} }.
\end{gather}
\end{subequations}
$\frm{W}$ is the wind frame, $\airDensity$ is the air density, $\airSpeed{}$ is the airspeed, $S$ is the planform area of the wing surface, and $c$ and $b$ are the mean aerodynamic chord and the wing span\cite{aerodynamics:book:beard2012small}. $C_D$, $C_L$, $C_Y$, $C_l$, $C_m$, and $C_n$ are the aerodynamic coefficients,
$\alpha$ is the angle of attack, $\beta$ is the sideslip angle, $\reynolds$ is the Reynolds number, and $\mach$ is the Mach number\cite{drones:book:UAV:valavanis2015handbook}.

\section{Co-Design Methodology} \label{sec:codesign-framework}

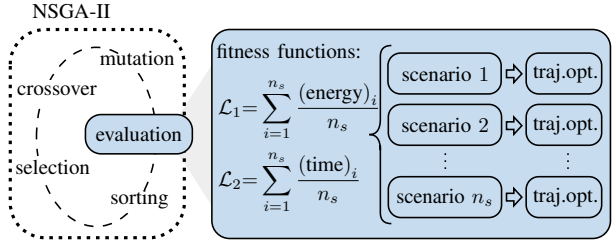
\begin{figure}[t]
    \centering
    \resizebox{1\columnwidth}{!}{
        \tikzset{every picture/.style={line width=0.75pt}} %

\begin{tikzpicture}[x=0.75pt,y=0.75pt,yscale=-1,xscale=1]

\draw  [draw opacity=0][fill={rgb, 255:red, 221; green, 221; blue, 221 }  ,fill opacity=0.48 ] (221.2,137.8) -- (255.2,85.8) -- (253.2,212.6) -- (220.4,161.8) -- cycle ;
\draw  [fill={rgb, 255:red, 199; green, 220; blue, 239 }  ,fill opacity=1 ] (242.2,94.56) .. controls (242.2,88.31) and (247.26,83.25) .. (253.51,83.25) -- (484.36,83.25) .. controls (490.61,83.25) and (495.67,88.31) .. (495.67,94.56) -- (495.67,204.95) .. controls (495.67,211.19) and (490.61,216.25) .. (484.36,216.25) -- (253.51,216.25) .. controls (247.26,216.25) and (242.2,211.19) .. (242.2,204.95) -- cycle ;
\draw  [color={rgb, 255:red, 0; green, 0; blue, 0 }  ,draw opacity=1 ][fill={rgb, 255:red, 136; green, 178; blue, 222 }  ,fill opacity=0 ][dash pattern={on 1.69pt off 2.76pt}][line width=1.5]  (114,105.38) .. controls (114,93.16) and (123.91,83.25) .. (136.13,83.25) -- (202.53,83.25) .. controls (214.76,83.25) and (224.67,93.16) .. (224.67,105.38) -- (224.67,193.62) .. controls (224.67,205.84) and (214.76,215.75) .. (202.53,215.75) -- (136.13,215.75) .. controls (123.91,215.75) and (114,205.84) .. (114,193.62) -- cycle ;
\draw  [dash pattern={on 0.84pt off 2.51pt}]  (390.67,162) -- (390.67,172) ;
\draw  [dash pattern={on 0.84pt off 2.51pt}]  (468.67,162) -- (468.67,172) ;
\draw  [dash pattern={on 4.5pt off 4.5pt}] (130.67,151.6) .. controls (130.67,120.12) and (147.69,94.6) .. (168.7,94.6) .. controls (189.71,94.6) and (206.73,120.12) .. (206.73,151.6) .. controls (206.73,183.08) and (189.71,208.6) .. (168.7,208.6) .. controls (147.69,208.6) and (130.67,183.08) .. (130.67,151.6) -- cycle ;
\draw   (356.34,93.44) .. controls (351.67,93.44) and (349.34,95.77) .. (349.34,100.44) -- (349.34,140.11) .. controls (349.34,146.78) and (347.01,150.11) .. (342.34,150.11) .. controls (347.01,150.11) and (349.34,153.44) .. (349.34,160.11)(349.34,157.11) -- (349.34,199.78) .. controls (349.34,204.45) and (351.67,206.78) .. (356.34,206.78) ;
\draw   (429.57,187.5) -- (434.22,187.5) -- (434.22,184.4) -- (439.33,189.92) -- (434.22,195.44) -- (434.22,192.34) -- (429.57,192.34) -- cycle ;
\draw   (429.57,142.01) -- (434.22,142.01) -- (434.22,138.91) -- (439.33,144.43) -- (434.22,149.95) -- (434.22,146.85) -- (429.57,146.85) -- cycle ;
\draw   (429.57,109.01) -- (434.22,109.01) -- (434.22,105.91) -- (439.33,111.43) -- (434.22,116.95) -- (434.22,113.85) -- (429.57,113.85) -- cycle ;
\draw  [draw opacity=0][fill={rgb, 255:red, 255; green, 255; blue, 255 }  ,fill opacity=1 ] (127.67,115.58) -- (143,115.58) -- (143,130.83) -- (127.67,130.83) -- cycle ;
\draw  [draw opacity=0][fill={rgb, 255:red, 255; green, 255; blue, 255 }  ,fill opacity=1 ] (126.67,163.92) -- (142,163.92) -- (142,179.17) -- (126.67,179.17) -- cycle ;
\draw  [draw opacity=0][fill={rgb, 255:red, 255; green, 255; blue, 255 }  ,fill opacity=1 ] (183.67,183.92) -- (199,183.92) -- (199,199.17) -- (183.67,199.17) -- cycle ;
\draw  [draw opacity=0][fill={rgb, 255:red, 255; green, 255; blue, 255 }  ,fill opacity=1 ] (183.33,95.58) -- (198.67,95.58) -- (198.67,110.83) -- (183.33,110.83) -- cycle ;

\draw  [fill={rgb, 255:red, 199; green, 220; blue, 239 }  ,fill opacity=1 ]  (161.79,147.38) .. controls (161.79,141.85) and (166.27,137.38) .. (171.79,137.38) -- (219.79,137.38) .. controls (225.31,137.38) and (229.79,141.85) .. (229.79,147.38) -- (229.79,152.38) .. controls (229.79,157.9) and (225.31,162.38) .. (219.79,162.38) -- (171.79,162.38) .. controls (166.27,162.38) and (161.79,157.9) .. (161.79,152.38) -- cycle  ;
\draw (195.79,149.88) node  [font=\normalsize] [align=left] {\begin{minipage}[lt]{43.8pt}\setlength\topsep{0pt}
\begin{center}
evaluation
\end{center}

\end{minipage}};
\draw (126.53,65.47) node [anchor=north west][inner sep=0.75pt]  [font=\normalsize,color={rgb, 255:red, 0; green, 0; blue, 0 }  ,opacity=1 ] [align=left] {NSGA-II};
\draw (243.67,90.67) node [anchor=north west][inner sep=0.75pt]  [font=\normalsize] [align=left] {fitness functions:};
\draw (244.67,113.11) node [anchor=north west][inner sep=0.75pt]  [font=\normalsize] [align=left] {$\displaystyle \cost{1} {=}\sum _{i=1}^{n_{s}}\frac{\left(\text{energy}\right)_{i}}{n_{s}}$};
\draw (244.67,159.11) node [anchor=north west][inner sep=0.75pt]  [font=\normalsize] [align=left] {$\displaystyle \cost{2} {=}\sum _{i=1}^{n_{s}}\frac{\left(\text{time}\right)_{i}}{n_{s}}$};
\draw    (354.17,106.25) .. controls (354.17,101.84) and (357.75,98.25) .. (362.17,98.25) -- (418.17,98.25) .. controls (422.59,98.25) and (426.17,101.84) .. (426.17,106.25) -- (426.17,116.25) .. controls (426.17,120.67) and (422.59,124.25) .. (418.17,124.25) -- (362.17,124.25) .. controls (357.75,124.25) and (354.17,120.67) .. (354.17,116.25) -- cycle  ;
\draw (390.17,111.25) node  [font=\normalsize] [align=left] {\begin{minipage}[lt]{46.01pt}\setlength\topsep{0pt}
\begin{center}
scenario $\displaystyle 1$
\end{center}

\end{minipage}};
\draw    (442.08,106.25) .. controls (442.08,101.84) and (445.67,98.25) .. (450.08,98.25) -- (482.08,98.25) .. controls (486.5,98.25) and (490.08,101.84) .. (490.08,106.25) -- (490.08,116.25) .. controls (490.08,120.67) and (486.5,124.25) .. (482.08,124.25) -- (450.08,124.25) .. controls (445.67,124.25) and (442.08,120.67) .. (442.08,116.25) -- cycle  ;
\draw (466.08,111.25) node  [font=\normalsize] [align=left] {\begin{minipage}[lt]{29.6pt}\setlength\topsep{0pt}
\begin{center}
traj.opt.
\end{center}

\end{minipage}};
\draw    (354.42,138.92) .. controls (354.42,134.5) and (358,130.92) .. (362.42,130.92) -- (418.42,130.92) .. controls (422.84,130.92) and (426.42,134.5) .. (426.42,138.92) -- (426.42,148.92) .. controls (426.42,153.34) and (422.84,156.92) .. (418.42,156.92) -- (362.42,156.92) .. controls (358,156.92) and (354.42,153.34) .. (354.42,148.92) -- cycle  ;
\draw (390.42,143.92) node  [font=\normalsize] [align=left] {\begin{minipage}[lt]{46.35pt}\setlength\topsep{0pt}
\begin{center}
scenario $\displaystyle 2$
\end{center}

\end{minipage}};
\draw    (442.08,138.92) .. controls (442.08,134.5) and (445.67,130.92) .. (450.08,130.92) -- (482.08,130.92) .. controls (486.5,130.92) and (490.08,134.5) .. (490.08,138.92) -- (490.08,148.92) .. controls (490.08,153.34) and (486.5,156.92) .. (482.08,156.92) -- (450.08,156.92) .. controls (445.67,156.92) and (442.08,153.34) .. (442.08,148.92) -- cycle  ;
\draw (466.08,143.92) node  [font=\normalsize] [align=left] {\begin{minipage}[lt]{29.6pt}\setlength\topsep{0pt}
\begin{center}
traj.opt.
\end{center}

\end{minipage}};
\draw    (354.42,184.92) .. controls (354.42,180.5) and (358,176.92) .. (362.42,176.92) -- (418.42,176.92) .. controls (422.84,176.92) and (426.42,180.5) .. (426.42,184.92) -- (426.42,194.92) .. controls (426.42,199.34) and (422.84,202.92) .. (418.42,202.92) -- (362.42,202.92) .. controls (358,202.92) and (354.42,199.34) .. (354.42,194.92) -- cycle  ;
\draw (390.42,189.92) node  [font=\normalsize] [align=left] {\begin{minipage}[lt]{46.35pt}\setlength\topsep{0pt}
\begin{center}
scenario $\displaystyle n_{s}$
\end{center}

\end{minipage}};
\draw    (442.08,184.92) .. controls (442.08,180.5) and (445.67,176.92) .. (450.08,176.92) -- (482.08,176.92) .. controls (486.5,176.92) and (490.08,180.5) .. (490.08,184.92) -- (490.08,194.92) .. controls (490.08,199.34) and (486.5,202.92) .. (482.08,202.92) -- (450.08,202.92) .. controls (445.67,202.92) and (442.08,199.34) .. (442.08,194.92) -- cycle  ;
\draw (466.08,189.92) node  [font=\normalsize] [align=left] {\begin{minipage}[lt]{29.6pt}\setlength\topsep{0pt}
\begin{center}
traj.opt.
\end{center}

\end{minipage}};
\draw (196.18,191.03) node  [font=\normalsize] [align=left] {\begin{minipage}[lt]{52.25pt}\setlength\topsep{0pt}
\begin{center}
sorting
\end{center}

\end{minipage}};
\draw (141,170.61) node  [font=\normalsize] [align=left] {\begin{minipage}[lt]{52.25pt}\setlength\topsep{0pt}
\begin{center}
selection
\end{center}

\end{minipage}};
\draw (143.6,122.14) node  [font=\normalsize] [align=left] {\begin{minipage}[lt]{52.25pt}\setlength\topsep{0pt}
\begin{center}
crossover
\end{center}

\end{minipage}};
\draw (194.75,101.23) node  [font=\normalsize] [align=left] {\begin{minipage}[lt]{52.25pt}\setlength\topsep{0pt}
\begin{center}
mutation
\end{center}

\end{minipage}};

\end{tikzpicture}
    }
	\caption{\justifying~Multi-objective optimisation of drone morphology to reduce energy consumption and mission completion time in diverse flight environments. The multi-objective assessment (fitness) of each individual topology involves $n_s$ trajectory optimisations.\looseness=-1}
    \label{fig:framework_nsga}
\end{figure}

The co-design method -- illustrated in \cref{fig:framework_nsga} -- identifies the conceptual design and the control strategy of morphing drones considering energy consumption, agility, and hardware limitations.
The design space is investigated with NSGA-II\cite{opti:EA:NSGA2:deb2002fast}.
Each drone is evaluated with a \textit{fitness functions} that measures the drone's ability to succeed in a set of $n_s$ scenarios.
We model the drone as a parametric multibody system subject to aerodynamics.
A scenario is an environment with obstacles where the drone, given some initial condition, is asked to complete a series of checkpoints using trajectory optimisation.\looseness=-1

The co-design parameters -- depicted in \cref{back:fig:platform} --  include:
$i\text{-}ii)$ wing chord and span size;
$iii\text{-}iv)$ wing vertical and horizontal location;
$v)$ wing static orientations $\rotmW{0}$;
$vi)$ kinematic chain of the \change{morphing mechanism};
$vii)$ servomotor models;
$viii)$ propulsion unit characteristics; and
$ix)$ controller weight $\psi$ -- discussed in \cref{subse:cost-function}.
The wing static orientation $\rotmW{0}$ is the wing's orientation w.r.t. the fuselage when the joints are in the rest configuration, namely $\jointPosRobot{=}\zero{}$. \change{From rotation} $\rotmW{0}$\change{, it is possible to identify the dihedral, incidence, and sweep in the rest configuration.}
The kinematic chain of the \change{morphing mechanism} is the description of the multi-body tree.
Specifically, the co-design method selects the number of joints interconnecting the fuselage and the wings, and it also identifies how these joints should be assembled, i.e., which axis of rotation must be actuated.
The co-design method selects the servomotor models to actuate each joint and identifies the propulsion unit selecting the optimal motor-propeller combination.
The co-design parameters are represented by an array of real values creating the \textit{chromosome}.
The fitness functions to be minimised are:\looseness=-1
\begin{equation*}
    \cost{1} = \sum_{i=1}^{n_s}\frac{ \brRound{ \text{energy}}_i}{n_s}, \quad
    \cost{2} = \sum_{i=1}^{n_s}\frac{ \brRound{ \text{time}}_i}{n_s}.
\end{equation*}
If the $i\text{-th}$ scenario is unfeasible, we assign $\brRound{ \text{energy}}_i$ and $\brRound{ \text{time}}_i$ a value of $10^6$.

\section{Morphing Drone Modelling} \label{sec:modelling}

The prediction of the drone behaviour relies on a mathematical model.
In detail, the drones are modelled as floating multibody systems subject to significant aerodynamic forces that cannot be ignored.
From the mechanical standpoint, the system comprises three primary links: the \textit{fuselage}, the \textit{left-wing}, and the \textit{right-wing}, all assumed to be rigid.
This section discusses joints, propulsion units, and aerodynamic modelling.\looseness=-1

\subsection{Mechanical Joints}
The \change{morphing mechanism interposed between fuselage and wing is composed} by $n$ one-DoF pin joints, each actuated with revolute DC servomotors with an integrated reduction system that enables a direct connection between the actuation and the links, eliminating the need for an additional reducer.
We use the same joint configuration to connect both the left wing and right wing to the fuselage to have a symmetric design. The system comprises $2n$ pin joints and $2n$ servomotors.
For simplicity, in the remainder of the paper, we use the word \textit{joints} to refer to all $2n$ pin joints which are actuated with servomotors.
The power consumption of a single revolute joint is estimated with the model \cite[Ch. 4.7]{fadini2023co}:\looseness=-1
\begin{equation} \label{eq:servo-motor:power-consumption}
    \power_{s} \hspace{-0.5mm} \brRound{ \dot{s}, \tau } = \dot{s} \tau {+} \dot{s} \tau_f {+} R k_v^2 \tau^2 + R k_v^2 \tau_f^2.
\end{equation}
$R$ is the electric motor resistance, $k_v$ is the motor constant velocity, $\dot{s}$ and $\tau$ are the joint velocity and torque. $\tau_f$ \change{is the friction torque due to viscous damping without Coulomb effects.}\looseness=-1

\subsection{Propulsion Unit}

The propulsion unit is implemented with an electric propeller placed on the fuselage nose, providing a positive force on the $\versorW{x}{\frm{B}}$ axis. 
\change{The propeller reaction torque acts on} $\versorW{x}{\frm{B}}$\change{, and its magnitude is modelled proportional to thrust}~$u$ \change{by a constant} $k_u$ \change{evaluated experimentally. We neglect gyroscopic and aerodynamic effects.}
The power consumption is modelled as\looseness=-1
\begin{equation} \label{eq:propeller:power-consumption}
    \power_{p} \hspace{-0.5mm} \brRound{u} = \xi_{0} + \xi_{1} u  + \xi_{2} u^2,
\end{equation}
where $\xi_{0}$, $\xi_{1}$, $\xi_{2}$ are constants evaluated experimentally. %

\subsection{Aerodynamics Forces} \label{subsec:aerodynamic-forces}

The aerodynamic forces (and moments) acting on a multibody system are dependent on $\alpha$, $\beta$, $\mach$, $\reynolds$, and $\jointPosRobot$. 
To simplify the aerodynamic analysis, we make these assumptions.
\looseness=-1
\begin{assumption} \label{ass:aero:mach}
    The $\mach$ effect on the aerodynamic forces and moments is negligible in our flight regime $\brRound{{\mach} {\ll} {1}}$, and the flow can be considered incompressible.
\end{assumption}
\begin{assumption} \label{ass:aero:multibody}
    The aerodynamic forces and moments acting on the multibody system are approximated using the superposition principle, wherein interactions among bodies are neglected. The total forces and moments are calculated as the sum of forces and moments acting on each body.
\end{assumption}
\begin{assumption} \label{ass:aero:ang-vel}
    The effects of rotational and unsteady motions of the platform on the surrounding airflow are negligible.
\end{assumption}
Assumption \ref{ass:aero:multibody} enables us to decompose the platform into three rigid bodies
and evaluate the aerodynamic forces acting on each body independently of the joint configuration $\jointPosRobot$. 
For the generic body $i$, the forces and moments can be modelled with \eqref{eq:aero:force-moments}, and act on a body frame $\frm{A}_i$ (see \cref{back:fig:platform}).
Assumption \ref{ass:aero:multibody} significantly decreases the number of required simulations or experiments.
For instance, if our platform implements 3-DoF joints between fuselage and wings, without assumption \ref{ass:aero:multibody}, we would need to perform $10^9$ analysis\footnote{Varying $\alpha {=} \brSquare{-10{:}2{:}10} \unit{\degree}$, $\beta{=}\brSquare{0{:}5{:}90} \unit{\degree}$, $\reynolds {=} \brSquare{1.2{:}0.8{:}3.6}{\times}10^5$ %
, and we change $s{=}\brSquare{-30{:}5{:}30}\unit{\degree}$ for all joints.}. 
Thanks to assumption \ref{ass:aero:multibody}, the number of simulations drops to $2500$ for each body\footnote{Varying $\alpha {=} \brSquare{-10{:}2{:}10} \unit{\degree}$, $\beta{=}\brSquare{0{:}5{:}90}\unit{\degree}$, and $\reynolds {=} \brSquare{1.2{:}0.8{:}3.6}{\times}10^5$.}\looseness=-1

\subsubsection{Aerodynamic Coefficients Identification} \label{ssubsec:aero-model-id}

we conducted aerodynamic simulations for all bodies using the \textit{$3$D uniform triangle panel Galerkin method} with the software \texttt{flow5}\cite{flow5}.
The fuselage was represented with a \textsc{naca0014} profile with a tail consisting of a horizontal and vertical stabiliser.
The left and right wings were represented by a \textsc{naca0009} profile with a constant chord. To capture differences in the wing's aspect ratio, defined as $b^2/S$, we performed separate aerodynamic modelling for each aspect ratio.
The aerodynamic simulations were performed in steady state configuration according to assumption \ref{ass:aero:ang-vel}, within the following parameter space:
$i)$ $\alpha {=} \brSquare{{-}10{:}2{:}10} \unit{\degree}$, $\beta{=}\brSquare{{-}30{:}2{:}30} \unit{\degree}$, $\reynolds {=} \brSquare{2{:}1{:}15}{\times}10^5$ for the fuselage; and
$ii)$ $\alpha {=} \brSquare{{-}10{:}2{:}10} \unit{\degree}$, $\beta{=}\brSquare{{-}130{:}2{:}130} \unit{\degree}$, $\reynolds {=} \brSquare{0.8{:}0.4{:}6.0}{\times}10^5$ for the wings.
The ranges of the angle of attack $\alpha$, sideslip angle $\beta$, and Reynolds number $\reynolds$ were selected based on the operating working conditions and the reliability range of the simulation software.\looseness=-1

\begin{figure}[t]
        \begin{subfigure}[b]{0.5\columnwidth}
            \centering
            \includegraphics[width=\textwidth, trim=0cm 0.3cm 0cm 0.75cm,clip]{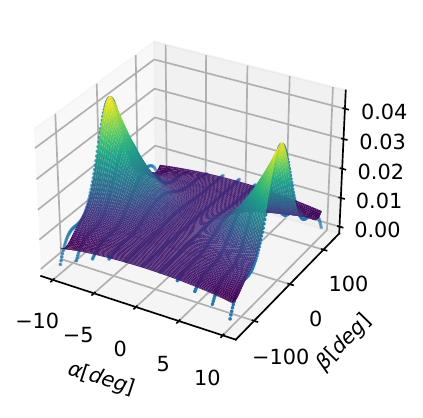}
            \caption{$C_D$}
            \label{fig:aero_coeff::CD}
        \end{subfigure}%
        \begin{subfigure}[b]{0.5\columnwidth}
            \centering
            \includegraphics[width=\textwidth, trim=0cm 0.3cm 0cm 0.75cm,clip]{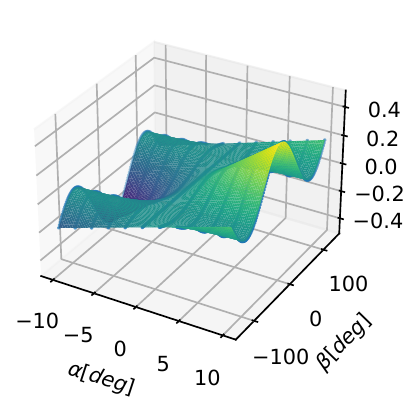}
            \caption{$C_L$}
            \label{fig:aero_coeff::CL}
        \end{subfigure}
        \caption{\justifying~Wing aerodynamic coefficients (airfoil \textsc{naca0009}, taper ratio $1.0$, aspect ratio $2.0$, ${\reynolds} {=} 2{\times}10^5$). The blue dots are the \texttt{flow5} data. The surfaces represent the analytical functions that fit data.
        \looseness=-1
        }
        \label{fig:aero_coeff}
\end{figure}

We used a linear regression approach to derive an analytical function that fits the simulation data. We chose a candidate function consisting of the sum of sinusoidal terms and solved the regression problem using Lasso regularisation%
~\cite{opti:convex:tibshirani1996regression}.
The results for a wing are shown in \cref{fig:aero_coeff}. \looseness=-1

\subsection{Equations of Motion}

The morphing drone dynamics are modelled with \eqref{eq:back:equation_motion}, where \looseness=-1
\begin{equation} \label{eq:mod:force}
    \boldsymbol{\mathrm{f}}^{\star} =
        \sum_{i=1}^{3} \jacobianRobot{\tr}{\frm{A}_i} \qvar \bmat{\forceBody{}{\boldsymbol{A}_i}  \\ \torqueBody{}{\boldsymbol{A}_i}} + 
        \jacobianRobot{\tr}{\frm{B}} \qvar
        \bmat{ \versorW{x}{\frm{B}} \qvar \\ k_u \, \versorW{x}{\frm{B}} \qvar } u.
\end{equation}
The aerodynamic forces and moments are evaluated with \eqref{eq:aero:force-moments} with the aerodynamic coefficients identified as in sec.~\ref{ssubsec:aero-model-id}.\looseness=-1

\section{Trajectory Optimisation} \label{sec:traj-opt}

The co-design method assesses drones' ability in complete scenarios.
The trajectory optimisation evaluates the desired \change{time-varying} \textit{actuation inputs} $\brRound{\jointTorRobot,\propThrustRobot}$ to navigate the drone in a given scenario being optimal according to a user \textit{metric}. 
The scenarios we consider involve reaching target locations with a desired attitude while avoiding obstacles.
The possible metrics that can be minimised include energy consumption and the time to complete the scenario.
The trajectory optimisation is transcribed using a direct multiple-shooting method, discretised in $N$ \textit{knots}, and formulated as an optimisation problem with cost function and constraints.

\subsection{Decision Variables}

The decision variables are:
joint configurations and their time derivatives $\jointPosRobot$, $\jointVelRobot$, $\jointAccRobot$; 
joint torques  $\jointTorRobot$, $\jointDotTorRobot$;
base position $\posW{\frmBase}$, $\linVelW{\frmBase}$, $\linAccW{\frmBase}$;
base attitude $\quatW{\frmBase}$, $\angVelW{\frmBase}$,  $\angAccW{\frmBase}$;
propeller thrust $\propThrustRobot$, $\propDotThrustRobot$; and
time interval between knots $\Delta t$.\looseness=-1

\subsection{Constraints}

\subsubsection{Initial Conditions}
we specify in the initial knot %
\begin{subequations} \label{eq:constraints:initial-condition}
\begin{gather}
    \posW{\frmBase} {\brSquare{0}} {=} \posW{\frmBase 0},       \hspace{2.9pt}
    \quatW{\frmBase} {\brSquare{0}} {=} \quatW{\frmBase 0},     \hspace{2.9pt}
    \linVelW{\frmBase} {\brSquare{0}} {=} \linVelW{\frmBase 0}, \hspace{2.9pt}
    \angVelW{\frmBase} {\brSquare{0}} {=} \angVelW{\frmBase 0}, \\
    \jointPosRobot {\brSquare{0}} {=} \jointPosRobot_0,         \hspace{4pt},
    \jointVelRobot {\brSquare{0}} {=} \jointVelRobot_0,         \hspace{4pt},
    \jointAccRobot {\brSquare{0}} {=} \jointAccRobot_0.
\end{gather}
\end{subequations}

\subsubsection{Hardware and Physical Limits}
we bound the decision variable for all the $N$ knots. For simplicity, the next constraints are expressed w.r.t. the generic knot $k$. %
For the joints and the propeller, we require:
\begin{subequations} \label{eq:constraints:box:joints_prop}
\begin{gather}
    \jointPosRobot_{\text{min}} {\le} \jointPosRobot {\brSquare{k}} {\le} \jointPosRobot_{\text{max}},        \hspace{4pt}
    \jointVelRobot_{\text{min}} {\le} \jointVelRobot {\brSquare{k}} {\le} \jointVelRobot_{\text{max}},        \hspace{4pt}
    \jointAccRobot_{\text{min}} {\le} \jointAccRobot {\brSquare{k}} {\le} \jointAccRobot_{\text{max}},        \\
    \jointTorRobot_{\text{min}} {\le} \jointTorRobot {\brSquare{k}} {\le} \jointTorRobot_{\text{max}},         \hspace{4pt}
    \propThrustRobot_{\text{min}} {\le} \propThrustRobot {\brSquare{k}} {\le} \propThrustRobot_{\text{max}}.
\end{gather}
\end{subequations}
To ensure the continuity of the control inputs, we bound:
\begin{equation} \label{eq:constraints:box:continuity}
    \jointDotTorRobot_{\text{min}} \le \jointDotTorRobot {\brSquare{k}} \le \jointDotTorRobot_{\text{max}}, \hspace{4pt}
    \propDotThrustRobot_{\text{min}} \le \propDotThrustRobot {\brSquare{k}} \le \propDotThrustRobot_{\text{max}}.
\end{equation}
The aerodynamic angles are constrained to remain within the reliable range of the model discussed in \cref{subsec:aerodynamic-forces}:
\begin{equation} \label{eq:constraints:box:aero-angles}
    \alpha_{\text{min}} \le \alpha {\brSquare{k}} \le \alpha_{\text{max}}, \hspace{4pt}
    \beta_{\text{min}} \le \beta {\brSquare{k}} \le \beta_{\text{max}}.
\end{equation}
Lastly, we constrain the time increment to be positive and lower than a threshold to prevent integration error, as:
\begin{equation} \label{eq:constraints:box:time}
    0 \le \Delta t {\brSquare{k}} \le \Delta t_{\text{max}}.
\end{equation}

\subsubsection{Obstacles Avoidance}

we require a positive distance between the drone and obstacles. 
We model the obstacles with $n_{\text{obs}}$ primitive shapes describable by a set $\mathcal{S}_{\mathcal{O}}$.
We identify $n_{\text{f}}$ significant drone points. %
For any point $\text{f}$ we must ensure\looseness=-1
\begin{equation} \label{eq:constraints:obstacle:sphere}
    \posW{\text{f}} {\brSquare{k}} {=} \FK{\posW{\text{f}}} {\brRound{\configurationRobot {\brSquare{k}} } }  \notin \mathcal{S}_{\mathcal{O}} \quad {\forall} k {\in} \brSquare{0{,} {...} {N}}.
\end{equation}

\subsubsection{Checkpoints}

we ask the drone to complete $n_{\text{cp}}$ checkpoints. 
A checkpoint $P$ is defined as $\mathcal{S}_P{=}\brRound{\posW{P}{,}\quatW{P}{,}\linVelW{P}{,}\angVelW{P}}$ 
in which the drone configuration should lie at the knot $k_P$, i.e.,\looseness=-1
\begin{equation} \label{eq:constraints:tasks}
    \brRound{
    \posW{\frmBase} {\brSquare{k_P}},
    \quatW{\frmBase} {\brSquare{k_P}},
    \linVelW{\frmBase} {\brSquare{k_P}},
    \angVelW{\frmBase} {\brSquare{k_P}}}
    \in 
    \mathcal{S}_P.
\end{equation}
The timing of each checkpoint is not defined by selecting the knot because $\Delta t$ is a decision variable.

\subsubsection{Decision Variables Integration}
we integrate the time-varying optimisation variables $\jointPosRobot$, $\jointVelRobot$, $\jointTorRobot$, $\propThrustRobot$, $\posW{\frmBase}$, $\linVelW{\frmBase}$, and $\angVelW{\frmBase}$ using the \textit{backward Euler method}. %

For $\quatW{\frmBase}$, we require
\begin{equation} \label{eq:constraints:integration:quat}
    \quatW{\frmBase} {\brSquare{k{+}1}} {=} \text{Exp}\brRound{\angVelW{} {\brSquare{k{+}1}} \Delta t {\brSquare{k}}}\quatW{\frmBase} {\brSquare{k}}. \\   
\end{equation}
$\text{Exp}\brRound{\cdot}$ implements the exponential operator which maps a $3$D angular velocity vector to a unit quaternion~\cite{control:lietheory:sola2018micro}. %

\subsubsection{System Dynamics}
we impose the dynamics that follow the modelling discussed in \cref{sec:modelling}, i.e.
\begin{equation}\label{eq:constraints:dynamics}
    \massMatrixRobot \brRound{\configurationRobot {\brSquare{k}}} \dotNuRobot {\brSquare{k}} + \biasVectorRobot \brRound{\configurationRobot {\brSquare{k}}, \nuRobot {\brSquare{k}}} = \bmat{\zero{6\times1} \\ \jointTorRobot {\brSquare{k}}} + \boldsymbol{\mathrm{f}}^{\star}
\end{equation}
for $k \in \brSquare{0, ... N}$. $\boldsymbol{\mathrm{f}}^{\star}$ is computed with \eqref{eq:mod:force}. 
$\alpha$, $\beta$, $\reynolds$, and $\airSpeed{}$
are computed for each body exploiting the kinematic chain.

\subsection{Cost Function} \label{subse:cost-function}
The cost function minimises a combination of time and energy consumption regulated by the weight $\psi$, namely:
\begin{equation} \label{eq:cost-function}
    \cost{} = \psi \sum_{k}^{N} \Delta t + \sum_{k}^{N} \brSquare{ \power_{p} \brRound{\propThrustRobot} + \sum_{j}^{n_{\text{j}}} \power_{s_j} \brRound{ \jointVelRobot, \jointTorRobot } } \Delta t.
\end{equation}

\subsection{Optimisation Problem}
The complete optimisation problem is formulated as
\begin{equation} \label{eq:optimisation-problem}
\setlength{\jot}{-1pt}
\begin{aligned}
    \text{min}_{\boldsymbol{\chi}} \quad & \psi \cdot \text{time} +   \text{energy} \quad & \text{eq. \eqref{eq:cost-function}} \\
    \text{s.t.} \quad & \text{initial conditions} & \text{eq. \eqref{eq:constraints:initial-condition}}\\
        & \text{hardware \& physical limits} & \text{eqs. \eqref{eq:constraints:box:joints_prop} to \eqref{eq:constraints:box:time}}  \\
        & \text{obstacles avoidance} & \text{eq. \eqref{eq:constraints:obstacle:sphere}}\\
        & \text{checkpoints} & \text{eq. \eqref{eq:constraints:tasks}} \\
        & \text{integration} & \text{eq. \eqref{eq:constraints:integration:quat}} \\
        & \text{system dynamics} & \text{eq. \eqref{eq:constraints:dynamics}}
\end{aligned}
\end{equation}
where $\boldsymbol{\chi}$ collects all the decision variables for all knots.

\section{Results} \label{sec:result}

Here, we test the method by exploring co-design for an agile drone capable of complex manoeuvres. Then, we validate it by comparing the performance of the co-designed morphing drones with a commercial fixed-wing drone.
\change{The code to reproduce the results is available online}\cite{repository}.\looseness=-1

\subsection{Co-Design Method} \label{subse:results:codesign}

The assessment of the co-design method requires the development of a framework. 
Our framework %
\url{https://bit.ly/icra2024}, 
is designed to enable users to: $i)$ define scenarios; $ii)$ modify trajectory optimisation formulation; and $iii)$ provide preferred aerodynamic models.
We implemented the trajectory optimisation problem using \texttt{CasADi} and solved using \texttt{Ipopt}~\cite{opti:ipopt:ipopt2006}, with the \texttt{ma27} linear solver~\cite{opti:hsl:website}.
The multibody system modelling is implemented using the \texttt{ADAM} library~\cite{Lerario2022}. %
NSGA-II is implemented using \texttt{DEAP} library~\cite{opti:deap:fortin2012deap}. The evolutionary process is characterised by a population of $100$ individuals, single-point crossover ($p{=}90{\%}$), random mutation ($p{=}6{\%}$), and a stop criteria after $100$ generations.\looseness=-1

\begin{figure*}[t]
        \centering
        \includegraphics[width=0.953\textwidth, trim=0cm 0cm 0cm 0.2cm,clip]{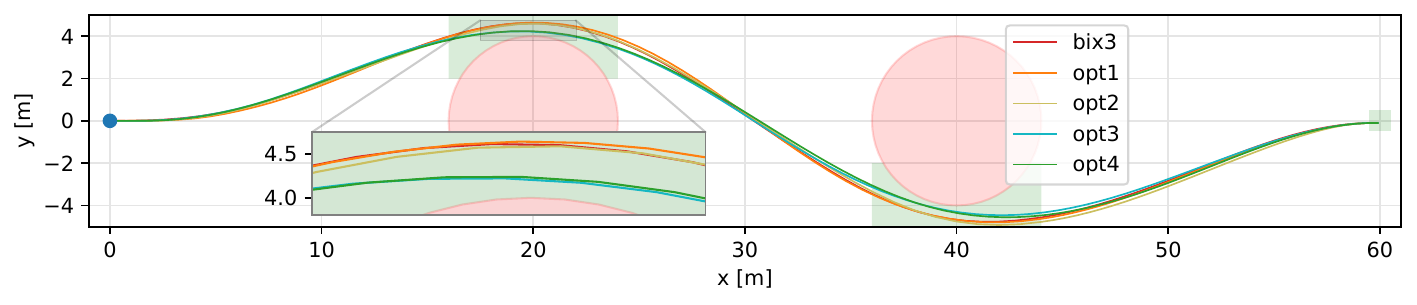}
        \captionsetup{skip=-2pt}
        \caption{\justifying~Trajectories of the \textsl{opt} drones. The red areas represent cylindrical obstacles, while the green areas represent checkpoints. %
        The drones follow similar paths except near obstacles, where the more agile solutions approach with lower clearance and higher speed (see \cref{fig:trajs_speed_x}).\looseness=-1}
        \label{fig:trajs_xy}
\end{figure*}

We investigated co-design of an agile drone capable of flying in airspace with obstacles that require complex manoeuvres, selecting five scenarios ($n_s{=}5$) for fitness evaluation.
Each scenario involves a slalom manoeuvre between two obstacles to evaluate the capability of changing directions.
The five scenarios share the same checkpoints and obstacles but vary in the drone's initial conditions, i.e., forward velocity ($8{\text{-}}12\unit{{\metre}{/}{\second}}$) and pitch orientation (${-}5{\text{-}}5\unit{\degree}$).
The obstacles are cylinders with infinite lengths and a ground plane.
The final target is $\posW{\frmBase}{=}\brRound{60,0,0}\unit{{\metre}}$ with $\linVelW{\frmBase}{=}\brRound{10,0,0}\unit{{\metre}{/}{\second}}$.
Two intermediate checkpoints are located near the obstacles.
The wind speed is $\brRound{-1,0,0}\unit{{\metre}{/}{\second}}$.
A scenario is depicted in \cref{fig:trajs_xy}.\looseness=-1

The co-design method was run nine times on a machine with \change{two AMD EPYC 7513 CPUs, utilising 100 cores}.
The average runtime for each execution was approximately $\qty{15.70}{\hour}$. Approximately $6400$ different individuals were analysed for each run, and $32000$ trajectory optimisation problems were solved.
In all tests, the fuselage and wings are assumed to be made of \textit{Expanded Polystyrene}. %
The fuselage has a length of $\qty{0.75}{\meter}$, a width of $\qty{0.1}{\meter}$, and it holds a payload of $\qty{0.35}{{\kilo}\gram}$ to account for the battery and the electronics.
The wing co-design parameters are listed in \cref{tab:wing-design-parameters}.
The method selects the propulsion unit from a database created based on online data\cite{rcbenchmark}. Similarly, the method chooses the servomotors to actuate the wings from a database of off-the-shelf Dynamixel servomotors.
\change{The off-the-shelf components selection influences the maximum servomotor torque} $\tau_{\text{max}}$, \change{maximum servomotor speed} $\omega_{\text{max}}$, \change{the maximum propeller thrust} $u_{\text{max}}$\change{, and their energy consumption.}
\looseness=-1

\begin{table}
    \centering
    \caption{\justifying~
Range of investigation for the wing co-design parameters. The wing span range is not predefined, in favor of the aspect ratio. %
    }
    \label{tab:wing-design-parameters}
        \begin{tabular}{l c|c|c|c}
        \toprule
        \multicolumn{2}{c|}{\textbf{Wing Design Parameter}} & \textbf{min} & \textbf{max} & \textbf{step} \\
        \midrule
        \rowcolor{gray!15}        
        chord size & ${\unit{\meter}}$ & $0.1$ &  $0.4$ & $0.05$ \\
        aspect ratio & $\cdot$ & $2$  &  $5$ &  $0.5$ \\
        \rowcolor{gray!15} 
        vertical location & ${\unit{\meter}}$ & $-0.03$  &  $0.03$ &  $0.01$ \\
        horizontal location & ${\unit{\meter}}$ & $-0.4$  &  $-0.1$ &  $0.05$ \\
        \rowcolor{gray!15} 
        wing static orientations & ${\unit{\degree}}$ & $-10$  &  $10$ &  $2$ \\
        \bottomrule
    \end{tabular}
\end{table}

\begin{figure*}[t]
    \hspace{1pt}
    \resizebox{\textwidth}{!}{\input{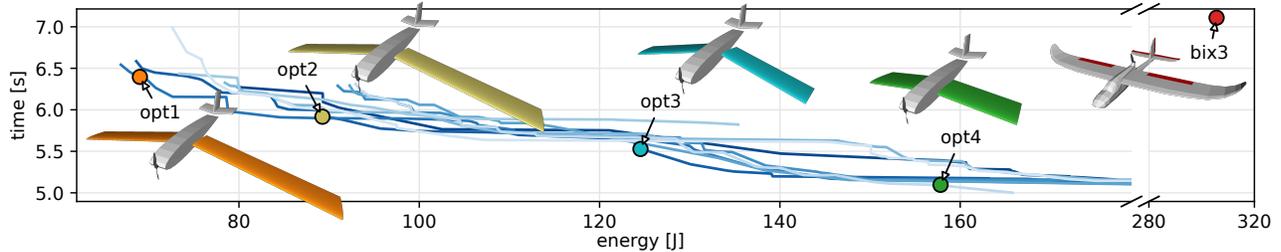}}
    \captionsetup{skip=-1pt}
    \caption{\justifying~Optimal Pareto front evaluated by the co-design optimisation method. 
    Each line represents the output of a single run.
    The markers identify the individuals analysed in \cref{fig:trajs_xy}, \cref{fig:trajs_speed_x}, and \cref{subse:results:codesign:validation}. \textit{bix3} is a commercial fixed-wing drone later used for validation.
    \change{The drones on the left side of the optimal Pareto fronts exhibit higher energy efficiency. Drones on the right bottom side show lower mission completion time.}\looseness=-1
    }
    \label{fig:pareto_front}
\end{figure*}

The Pareto fronts, evaluated by the co-design method in the \change{nine} runs, are shown in \cref{fig:pareto_front}.
The resulting drones share common characteristics: a negative static dihedral, positive wing static angle of attack, wings mounted in the upper part of the fuselage, and a chord length of $\qty{0.24}{{\metre}}$.
Agile drones tend to have a lower aspect ratio, a powerful propulsion unit, higher gain $\psi$, and are often equipped with three servomotors per wing.
Differently, energy-efficiency drones present opposite characteristics and are usually designed with two servomotors per wing (sweep \& incidence) or, in some cases, only one (incidence).
In terms of servomotor models, the dihedral motors are usually more powerful ($\tau_{\text{max}}{=}\qty{3.7}{{{\newton}}{{\metre}}}$) than sweep ($\tau_{\text{max}}{=}\qty{0.46}{{\newton}{\metre}}$) and incidence ($\tau_{\text{max}}{=}\qty{0.36}{{\newton}{\metre}}$).
The drone's weight is influenced by wing size, number of joints, and servomotor models. As a result, energy-efficient drones tend to have lower mass ($\qty{0.8}{{\kilo}\gram}$) than their agile counterparts ($\qty{1}{{\kilo}\gram}$).
\looseness=-1

\begin{figure}[t]
        \centering
        \includegraphics[width=\columnwidth]{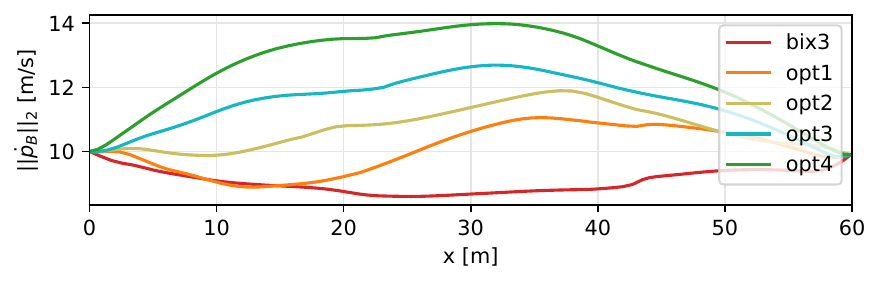}
        \captionsetup{skip=-3pt}
        \caption{\justifying~Velocity profiles of co-designed drones during slalom trajectory. 
        \textsl{opt3} and \textsl{opt4} achieve higher speeds \change{and complete the scenario in} $30{\%}$ and $39{\%}$ \change{faster than \textsl{bix3}, respectively.}
        \looseness=-1}
        \label{fig:trajs_speed_x}
\end{figure}

Figure \ref{fig:pareto_front} shows four co-designed drones labelled as \textsl{opt1}, \textsl{opt2}, \textsl{opt3}, and \textsl{opt4}.
\textsl{opt1} is equipped with a single joint per wing for actuating the incidence angle, wing aspect ratio  $4.5$, and has a propeller with $\propThrustRobot_{\text{max}}{=}\qty{4}{{\newton}}$.
\textsl{opt2} has two revolute joints to actuate sweep and incidence, wing aspect ratio  $4.5$, and $\propThrustRobot_{\text{max}}{=}\qty{4}{{\newton}}$. 
\textsl{opt3} incorporates three revolute joints to actuate incidence, sweep, and dihedral, wing aspect ratio  $3.0$, and $\propThrustRobot_{\text{max}}{=}\qty{4}{{\newton}}$. 
\textsl{opt4} has three revolute joints to actuate dihedral, sweep, and incidence, wing aspect ratio  $2$, and $\propThrustRobot_{\text{max}}{=}\qty{10}{{\newton}}$.
Figures \ref{fig:trajs_xy} and \ref{fig:trajs_speed_x} report the trajectories of the \textsl{opt} drones during a scenario with an initial forward velocity of $10\unit{{\metre}{/}{\second}}$ and zero initial pitch orientation.
\looseness=-1

\subsection{Co-Design Validation} \label{subse:results:codesign:validation}

In \cref{subse:results:codesign}, we tested the co-design method designing an agile drone capable of avoiding obstacles by studying $n_s{=}5$ scenarios; however, drones may operate in different conditions.
We now assess the capabilities of the optimal drones in environments different from the optimised ones.
In our analysis, we include the four \textsl{opt} drones represented in \cref{fig:pareto_front}; and the commercial drone \textit{H-King Bixler3} with fixed-wing and control surfaces which serves as a baseline as it is of similar weight and size as the \textsl{opt} drones. We refer to this drone as \textsl{bix3}. The aerodynamic model of \textsl{bix3} was obtained by combining the wind tunnel results with the data presented in~\cite{drones:bixler3:waldock2018learning}.
\textsl{bix3} has a propeller which can provide \change{maximum thrust} $\propThrustRobot_{\text{max}}{=}\qty{7}{{\newton}}$, a total wingspan of $\qty{1.55}{\metre}$, and weights $\qty{1012}{\gram}$ when equipped with sensors for autonomous flight\cite{wuest2022accurate}. 
\looseness=-1

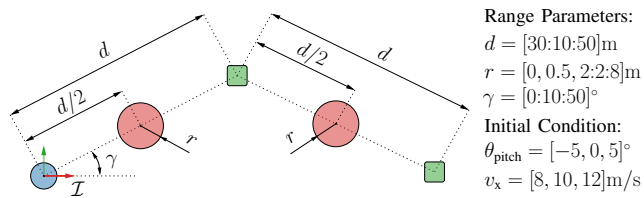
\begin{figure}[t]
    \centering
    \resizebox{\columnwidth}{!}{
        \tikzset{every picture/.style={line width=0.75pt}} %

\begin{tikzpicture}[x=0.75pt,y=0.75pt,yscale=-1,xscale=1]

\draw  [dash pattern={on 0.84pt off 2.51pt}]  (294,103) -- (83.3,212.3) ;
\draw  [dash pattern={on 0.84pt off 2.51pt}]  (294,103) -- (509,208) ;
\draw  [fill={rgb, 255:red, 214; green, 39; blue, 40 }  ,fill opacity=0.5 ] (376.5,155.5) .. controls (376.5,141.69) and (387.69,130.5) .. (401.5,130.5) .. controls (415.31,130.5) and (426.5,141.69) .. (426.5,155.5) .. controls (426.5,169.31) and (415.31,180.5) .. (401.5,180.5) .. controls (387.69,180.5) and (376.5,169.31) .. (376.5,155.5) -- cycle ;
\draw    (255.77,37.47) -- (48.63,144.93) ;
\draw [shift={(46.85,145.85)}, rotate = 332.58] [fill={rgb, 255:red, 0; green, 0; blue, 0 }  ][line width=0.08]  [draw opacity=0] (9.6,-2.4) -- (0,0) -- (9.6,2.4) -- cycle    ;
\draw [shift={(257.55,36.55)}, rotate = 152.58] [fill={rgb, 255:red, 0; green, 0; blue, 0 }  ][line width=0.08]  [draw opacity=0] (9.6,-2.4) -- (0,0) -- (9.6,2.4) -- cycle    ;
\draw  [dash pattern={on 0.84pt off 2.51pt}]  (46.85,145.85) -- (83.3,212.3) ;
\draw  [dash pattern={on 0.84pt off 2.51pt}]  (257.55,36.55) -- (294,103) ;
\draw    (333.3,36.38) -- (544.7,139.62) ;
\draw [shift={(546.5,140.5)}, rotate = 206.03] [fill={rgb, 255:red, 0; green, 0; blue, 0 }  ][line width=0.08]  [draw opacity=0] (9.6,-2.4) -- (0,0) -- (9.6,2.4) -- cycle    ;
\draw [shift={(331.5,35.5)}, rotate = 26.03] [fill={rgb, 255:red, 0; green, 0; blue, 0 }  ][line width=0.08]  [draw opacity=0] (9.6,-2.4) -- (0,0) -- (9.6,2.4) -- cycle    ;
\draw  [dash pattern={on 0.84pt off 2.51pt}]  (331.5,35.5) -- (294,103) ;
\draw  [dash pattern={on 0.84pt off 2.51pt}]  (546.5,140.5) -- (509,208) ;
\draw  [fill={rgb, 255:red, 214; green, 39; blue, 40 }  ,fill opacity=0.5 ] (163.65,157.65) .. controls (163.65,143.84) and (174.84,132.65) .. (188.65,132.65) .. controls (202.46,132.65) and (213.65,143.84) .. (213.65,157.65) .. controls (213.65,171.46) and (202.46,182.65) .. (188.65,182.65) .. controls (174.84,182.65) and (163.65,171.46) .. (163.65,157.65) -- cycle ;
\draw    (144.44,209.45) .. controls (143.38,196.04) and (143.25,197.33) .. (136.59,187.08) ;
\draw [shift={(135.6,185.55)}, rotate = 57.38] [fill={rgb, 255:red, 0; green, 0; blue, 0 }  ][line width=0.08]  [draw opacity=0] (8.4,-2.1) -- (0,0) -- (8.4,2.1) -- cycle    ;
\draw [shift={(144.6,211.55)}, rotate = 265.55] [fill={rgb, 255:red, 0; green, 0; blue, 0 }  ][line width=0.08]  [draw opacity=0] (8.4,-2.1) -- (0,0) -- (8.4,2.1) -- cycle    ;
\draw  [dash pattern={on 0.84pt off 2.51pt}]  (83.3,212.3) -- (183.3,212.3) ;
\draw  [fill={rgb, 255:red, 31; green, 119; blue, 180 }  ,fill opacity=0.5 ] (69,212.3) .. controls (69,204.4) and (75.4,198) .. (83.3,198) .. controls (91.2,198) and (97.6,204.4) .. (97.6,212.3) .. controls (97.6,220.2) and (91.2,226.6) .. (83.3,226.6) .. controls (75.4,226.6) and (69,220.2) .. (69,212.3) -- cycle ;
\draw [color={rgb, 255:red, 44; green, 160; blue, 44 }  ,draw opacity=0.5 ][line width=1.5]    (83.3,212.3) -- (83.3,182.85) ;
\draw [shift={(83.3,178.85)}, rotate = 90] [fill={rgb, 255:red, 44; green, 160; blue, 44 }  ,fill opacity=1 ][line width=0.08]  [draw opacity=0] (10.92,-2.73) -- (0,0) -- (10.92,2.73) -- cycle    ;
\draw [color={rgb, 255:red, 214; green, 39; blue, 40 }  ,draw opacity=1 ][line width=1.5]    (83.3,212.3) -- (113.6,212.3) ;
\draw [shift={(117.6,212.3)}, rotate = 180] [fill={rgb, 255:red, 214; green, 39; blue, 40 }  ,fill opacity=1 ][line width=0.08]  [draw opacity=0] (10.92,-2.73) -- (0,0) -- (10.92,2.73) -- cycle    ;
\draw  [fill={rgb, 255:red, 0; green, 0; blue, 0 }  ,fill opacity=1 ] (82.8,212.3) .. controls (82.8,212.02) and (83.02,211.8) .. (83.3,211.8) .. controls (83.58,211.8) and (83.8,212.02) .. (83.8,212.3) .. controls (83.8,212.58) and (83.58,212.8) .. (83.3,212.8) .. controls (83.02,212.8) and (82.8,212.58) .. (82.8,212.3) -- cycle ;
\draw  [fill={rgb, 255:red, 0; green, 0; blue, 0 }  ,fill opacity=1 ] (188.15,157.65) .. controls (188.15,157.37) and (188.37,157.15) .. (188.65,157.15) .. controls (188.93,157.15) and (189.15,157.37) .. (189.15,157.65) .. controls (189.15,157.93) and (188.93,158.15) .. (188.65,158.15) .. controls (188.37,158.15) and (188.15,157.93) .. (188.15,157.65) -- cycle ;
\draw    (188.65,157.65) -- (241.97,187.05) ;
\draw [shift={(209.27,169.02)}, rotate = 28.87] [fill={rgb, 255:red, 0; green, 0; blue, 0 }  ][line width=0.08]  [draw opacity=0] (8.4,-2.1) -- (0,0) -- (8.4,2.1) -- cycle    ;

\draw  [fill={rgb, 255:red, 0; green, 0; blue, 0 }  ,fill opacity=1 ] (401.7,155.04) .. controls (401.95,155.15) and (402.07,155.45) .. (401.96,155.7) .. controls (401.85,155.95) and (401.55,156.07) .. (401.3,155.96) .. controls (401.05,155.85) and (400.93,155.55) .. (401.04,155.3) .. controls (401.15,155.05) and (401.45,154.93) .. (401.7,155.04) -- cycle ;
\draw    (401.5,155.5) -- (352,188.33) ;
\draw [shift={(382.5,168.1)}, rotate = 146.44] [fill={rgb, 255:red, 0; green, 0; blue, 0 }  ][line width=0.08]  [draw opacity=0] (8.4,-2.1) -- (0,0) -- (8.4,2.1) -- cycle    ;
\draw  [dash pattern={on 0.84pt off 2.51pt}]  (168.37,120.06) -- (188.65,157.15) ;
\draw    (166.6,120.98) -- (64.8,173.79) ;
\draw [shift={(63.02,174.71)}, rotate = 332.58] [fill={rgb, 255:red, 0; green, 0; blue, 0 }  ][line width=0.08]  [draw opacity=0] (9.6,-2.4) -- (0,0) -- (9.6,2.4) -- cycle    ;
\draw [shift={(168.37,120.06)}, rotate = 152.58] [fill={rgb, 255:red, 0; green, 0; blue, 0 }  ][line width=0.08]  [draw opacity=0] (9.6,-2.4) -- (0,0) -- (9.6,2.4) -- cycle    ;
\draw  [dash pattern={on 0.84pt off 2.51pt}]  (422.2,118.8) -- (401.7,155.04) ;
\draw    (316.5,67.18) -- (420.4,117.92) ;
\draw [shift={(422.2,118.8)}, rotate = 206.03] [fill={rgb, 255:red, 0; green, 0; blue, 0 }  ][line width=0.08]  [draw opacity=0] (9.6,-2.4) -- (0,0) -- (9.6,2.4) -- cycle    ;
\draw [shift={(314.7,66.3)}, rotate = 26.03] [fill={rgb, 255:red, 0; green, 0; blue, 0 }  ][line width=0.08]  [draw opacity=0] (9.6,-2.4) -- (0,0) -- (9.6,2.4) -- cycle    ;
\draw  [fill={rgb, 255:red, 44; green, 160; blue, 44 }  ,fill opacity=0.5 ] (283.21,94.42) .. controls (283.21,93.11) and (284.28,92.05) .. (285.59,92.05) -- (302.41,92.05) .. controls (303.72,92.05) and (304.79,93.11) .. (304.79,94.42) -- (304.79,111.58) .. controls (304.79,112.89) and (303.72,113.95) .. (302.41,113.95) -- (285.59,113.95) .. controls (284.28,113.95) and (283.21,112.89) .. (283.21,111.58) -- cycle ;
\draw  [fill={rgb, 255:red, 44; green, 160; blue, 44 }  ,fill opacity=0.5 ] (498.21,199.42) .. controls (498.21,198.11) and (499.28,197.05) .. (500.59,197.05) -- (517.41,197.05) .. controls (518.72,197.05) and (519.79,198.11) .. (519.79,199.42) -- (519.79,216.58) .. controls (519.79,217.89) and (518.72,218.95) .. (517.41,218.95) -- (500.59,218.95) .. controls (499.28,218.95) and (498.21,217.89) .. (498.21,216.58) -- cycle ;

\draw (149.5,187.5) node [anchor=north west][inner sep=0.75pt]  [font=\LARGE] [align=left] {$\displaystyle \gamma $};
\draw (136.84,69.25) node [anchor=north west][inner sep=0.75pt]  [font=\LARGE,rotate=-334.14] [align=left] {$\displaystyle d$};
\draw (452.15,66.22) node [anchor=north west][inner sep=0.75pt]  [font=\LARGE,rotate=-26.19] [align=left] {$\displaystyle d$};
\draw (241.64,164.77) node [anchor=north west][inner sep=0.75pt]  [font=\LARGE,rotate=-28.99] [align=left] {$\displaystyle r$};
\draw (343.87,168.89) node [anchor=north west][inner sep=0.75pt]  [font=\LARGE,rotate=-323.92] [align=left] {$\displaystyle r$};
\draw (90.81,129.47) node [anchor=north west][inner sep=0.75pt]  [font=\LARGE,rotate=-332.41] [align=left] {$\displaystyle d/2$};
\draw (364.5,60.71) node [anchor=north west][inner sep=0.75pt]  [font=\LARGE,rotate=-25.47] [align=left] {$\displaystyle d/2$};
\draw (112,217.4) node [anchor=north west][inner sep=0.75pt]  [font=\LARGE]  {$\mathcal{I}$};
\draw (562,56.4) node [anchor=north west][inner sep=0.75pt]  [font=\LARGE]  {$d=[ 30{:}10{:}50]\unit{\metre}$};
\draw (560,114.4) node [anchor=north west][inner sep=0.75pt]  [font=\LARGE]  {$\gamma =[ 0{:}10{:}50]\unit{\degree}$};
\draw (562,86.4) node [anchor=north west][inner sep=0.75pt]  [font=\LARGE]  {$r=[ 0,0.5,2{:}2{:}8]\unit{\metre}$};
\draw (562,173.4) node [anchor=north west][inner sep=0.75pt]  [font=\LARGE]  {$\theta _{\text{pitch}} =[ {-}5,0,5]\unit{\degree}$};
\draw (562,204.4) node [anchor=north west][inner sep=0.75pt]  [font=\LARGE]  {$v_{\text{x}} =[ 8,10,12]\unit{\metre/\second}$};
\draw (562,148) node [anchor=north west][inner sep=0.75pt]  [font=\LARGE] [align=left] {Initial Condition:};
\draw (562,29) node [anchor=north west][inner sep=0.75pt]  [font=\LARGE] [align=left] {Range Parameters:};

\end{tikzpicture}
    }
	\caption{\justifying~Parametric scenario to validate the co-design method. The drone starts at the blue sphere and must reach two (green) checkpoints while avoiding obstacles (red spheres). \change{The scenarios with }$r=\qty{0}{\metre}$\change{ are without obstacles.}\looseness=-1}
    \label{fig:validation_scenario}
\end{figure}

\begin{figure}[t]
        \begin{subfigure}[b]{0.5\columnwidth}
            \includegraphics[width=\textwidth]{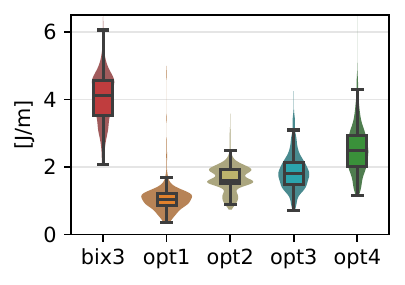}
            \caption{energy normalised \change{by} $d$}
            \label{fig:validation:energy}
        \end{subfigure}%
        \begin{subfigure}[b]{0.5\columnwidth}
            \includegraphics[width=\textwidth]{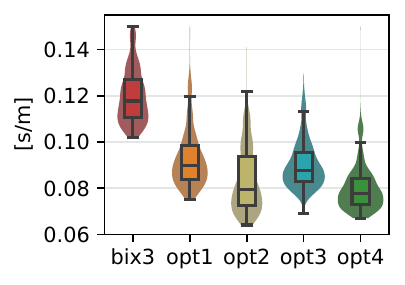}
            \caption{time normalised \change{by} $d$}
            \label{fig:validation:time}
        \end{subfigure}
        \caption{\justifying~Results of the validation of co-design methodology. Energy consumption and mission time are divided by length $d$ \change{-- defined in }\cref{fig:validation_scenario} \change{--} for comparing scenarios with different \change{mission distances}.\looseness=-1}
        \label{fig:validation}
\end{figure}

\textsl{bix3} and \textsl{opt} drones are evaluated in a common parametric scenario depicted in \cref{fig:validation_scenario}.
For each drone, we performed $972$ simulations by varying the parameters $d$, $\gamma$, $r$, $v_x$, and $\theta_{\text{p}}$ in all possible configurations.
$v_x$ is the forward velocity, and $\theta_{\text{p}}$ is the pitch orientation.
The results of the $972{\times}5$ simulations are shown in \cref{fig:validation}.
Co-designed drones exhibit a decrease in average energy consumption by $37{\text{-}}74{\%}$ and a decrease in time to complete the scenario by $22{\text{-}}33{\%}$ compared to \textsl{bix3}.
Therefore, the resulting \textsl{opt} drones outperform the commonly used commercial platform in mission time and energy efficiency in our tested scenarios, \change{including those without obstacles}.\looseness=-1

\section{Conclusions} \label{sec:conclusions}

Co-design methods can assist engineers in developing more efficient drones, but existing methods are inadequate for drones with morphing wings.
This paper addresses this gap by proposing a co-design methodology that identifies topology, actuation, morphing strategy, and controller parameters.
Our method relies on a parametric modelling approach, trajectory optimisation, and a multi-objective optimisation algorithm.\looseness=-1

The method was tested for the co-design of morphing topology and control in five airspace scenarios with checkpoints and obstacles, to minimise energy consumption and mission time.  However, the method could also be used with other objectives and in different scenarios.\looseness=-1

Morphing drones' aerodynamics were modelled with an approach valid at low angles of attack -- forcing us to limit the wing range of movements -- and we neglected the interactions between wings and fuselage to reduce computational complexities. Relaxing these assumptions would lead to a more realistic model that can unlock aggressive manoeuvres.\looseness=-1

Future work could include \change{the design of a physical prototype with the presented methodology}, the integration of a dynamic simulator to study wind disturbance rejection, and the enlargement of the design space to include propeller positioning, battery selection, and other fuselage and tail configurations. \change{Additionally, we could incorporate ailerons, elevators, and a rudder as co-design parameters, creating drones that combine conventional control surfaces with morphing wings, enhancing manoeuvrability or proposing energy-efficient solutions with a lower number of actuators.}  \looseness=-1

The topology and control co-design method described here could be used to assist aircraft engineers in the initial design phase of agile and energy-efficient morphing drones that \change{meet} mission-specific costs and constraints.\looseness=-1

\addtolength{\textheight}{-4cm}   %

\bibliographystyle{IEEEtran}
\bibliography{IEEEabrv,bibliography}

\end{document}